\documentclass[11pt,a4paper]{article}
\usepackage{times,latexsym}
\usepackage{url}
\usepackage[T1]{fontenc}

\usepackage[acceptedWithA]{tacl2021v1}

\usepackage{xspace,mfirstuc,tabulary}

\newif\iftaclinstructions
\taclinstructionsfalse %
\iftaclinstructions
\renewcommand{\confidential}{}
\renewcommand{\anonsubtext}{(No author info supplied here, for consistency with
TACL-submission anonymization requirements)}
\newcommand{\instr}
\fi

\iftaclpubformat %

\else

\fi

\usepackage[utf8]{inputenc}

\usepackage{microtype}

\usepackage{inconsolata}

\usepackage{multicol}
\usepackage{multirow}
\usepackage{graphicx}
\usepackage{booktabs}
\usepackage{xspace}

\newcommand{\mptinstruct}[0]{MPT-30B-Instruct\xspace}
\newcommand{\mpt}[0]{MPT-30B\xspace}
\newcommand{\longchat}[0]{LongChat-13B (16K)\xspace}
\newcommand{\gptturbo}[0]{GPT-3.5-Turbo\xspace}
\newcommand{\gptturboextended}[0]{GPT-3.5-Turbo (16K)\xspace}
\newcommand{\claude}[0]{Claude-1.3\xspace}
\newcommand{\claudeextended}[0]{Claude-1.3 (100K)\xspace}
\newcommand{\gptfour}[0]{GPT-4\xspace}
\newcommand{\flanultwo}[0]{Flan-UL2\xspace}
\newcommand{\flantfive}[0]{Flan-T5-XXL\xspace}
\newcommand{\emldisplay}[2]{\texttt{\href{mailto:#1}{#2}}}
\newcommand{\eml}[1]{\emldisplay{#1}{#1}}
\let\svthefootnote\thefootnote
\newcommand\blankfootnote[1]{%
  \let\thefootnote\relax\footnotetext{#1}%
  \let\thefootnote\svthefootnote%
}

\let\oldFootnote\footnote
\newcommand\nextToken\relax

\renewcommand\footnote[1]{%
    \oldFootnote{#1}\futurelet\nextToken\isFootnote}

\newcommand\isFootnote{%
    \ifx\footnote\nextToken\textsuperscript{,}\fi}

\newif\ifcomments
\commentstrue

\ifcomments
    \providecommand{\nfl}[1]{{\protect\color{Green}{[NFL: #1]}}}
    \providecommand{\ap}[1]{{\protect\color{Red}{[AP: #1]}}}
    \providecommand{\mb}[1]{{\protect\color{ProcessBlue}{[MB: #1]}}}
    \providecommand{\kl}[1]{{\protect\color{ProcessBlue}{[KL: #1]}}}
    \providecommand{\pl}[1]{{\protect\color{Red}{[PL: #1]}}}
\else
    \providecommand{\nfl}[1]{}
    \providecommand{\pl}[1]{}
    \providecommand{\kl}[1]{}
    \providecommand{\ap}[1]{}
    \providecommand{\mb}[1]{}
\fi

\newcommand\tab[1][0.7cm]{\hspace*{#1}}
\newcommand{\affa}{{$^{1}$}}
\newcommand{\affb}{{$^{2}$}}
\newcommand{\affc}{{$^{3}$}}

\title{Lost in the Middle: How Language Models Use Long Contexts}

\author{Nelson F. Liu$^{1*}$ \tab Kevin Lin\affb \tab John Hewitt\affa \tab Ashwin Paranjape\affc \\ \quad {\bf Michele Bevilacqua}\affc \tab  {\bf Fabio Petroni}\affc \tab {\bf Percy Liang}\affa \\
  \affa Stanford University \tab \affb University of California, Berkeley \tab \affc Samaya AI \\
  \eml{nfliu@cs.stanford.edu}
}

\begin{document}

\maketitle
\blankfootnote{\llap{\textsuperscript{*}}Work partially completed as an intern at Samaya AI.}
\begin{abstract}
While recent language models have the ability to take long contexts as input, relatively little is known about how well they \textit{use} longer context.
We analyze the performance of language models on two tasks that require identifying relevant information in their input contexts: multi-document question answering and key-value retrieval.
We find that performance can degrade significantly when changing the position of relevant information, indicating that current language models do not robustly make use of information in long input contexts.
In particular, we observe that performance is often highest when relevant information occurs at the beginning or end of the input context, and significantly degrades when models must access relevant information in the middle of long contexts, even for explicitly long-context models.
Our analysis provides a better understanding of how language models use their input context and provides new evaluation protocols for future long-context language models.
\end{abstract}

\section{Introduction}

\begin{figure}[t]
\centering
\includegraphics[width=\columnwidth]{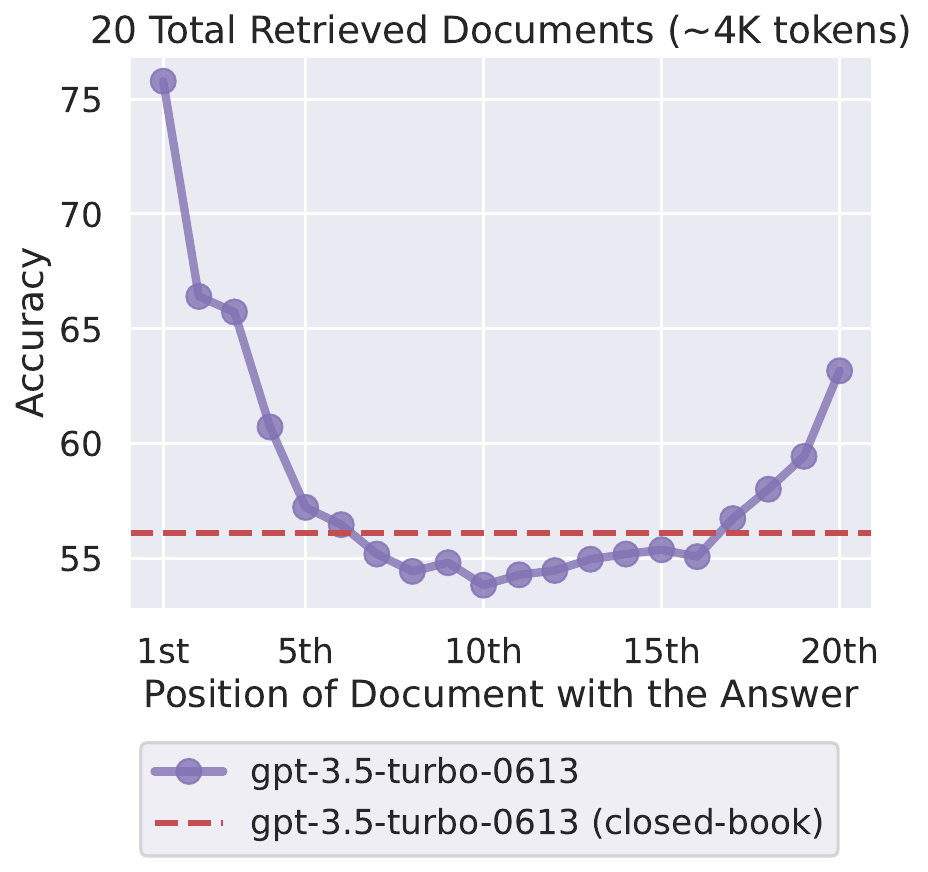}
\caption{Changing the location of relevant information (in this case, the position of the passage that answers an input question) within the language model's input context results in a U-shaped performance curve---models are better at using relevant information that occurs at the very beginning (primacy bias) or end of its input context (recency bias), and performance degrades significantly when models must access and use information located in the middle of its input context.}\label{fig:figure1}
\end{figure}

Language models have become an important and flexible building block in a variety of user-facing language technologies, including conversational interfaces, search and summarization, and collaborative writing \citep[\emph{inter alia}]{shuster2022blenderbot,thoppilan2022lamda,lee2022coauthor}.
These models perform downstream tasks primarily via prompting: all relevant task specification and data to process is formatted as a textual input context, and the model returns a generated text completion.
These input contexts can contain thousands of tokens, especially when language models are used to process long documents (e.g., legal or scientific documents, conversation histories, etc.) or when language models are augmented with external information (e.g., relevant documents from a search engine, database query results, etc; \citealp[\emph{inter alia}]{petroni2020context,ram2023incontext,shi2023replug,mallen2023trust,schick2023toolformer}).

Handling these use-cases requires language models to successfully operate over long sequences.
Existing language models are generally implemented with Transformers \citep{10.5555/3295222.3295349}, which require memory and compute that increases quadratically in sequence length.
As a result, Transformer language models were often trained with relatively small context windows (between 512-2048 tokens).
Recent improvements in hardware (e.g., faster GPUs with more memory) and algorithms \citep[\emph{inter alia}]{dai-etal-2019-transformer,dao2022flashattention,poli2023hyena,rubin2023longrange} have resulted in language models with larger context windows (e.g., 4096, 32K, and even 100K tokens), but it remains unclear how these extended-context language models make use of their input contexts when performing downstream tasks.

We empirically investigate this question via controlled experiments with a variety of state-of-the-art open (\mptinstruct, \longchat) and closed (OpenAI's \gptturbo and Anthropic's \claude) language models in settings that require accessing and using information within an input context.
In particular, our experiments make controlled changes to the input context size and the position of the relevant information within the input context and study their effects on language model performance.
If language models can robustly use information within long input contexts, then their performance should be \emph{minimally affected} by the position of the relevant information in the input context.

We first experiment with multi-document question answering, which requires models to reason over provided documents to find relevant information and use it to answer a given question; this task mimics the retrieval-augmented generation setup underlying many commercial generative search and question answering applications (e.g., Bing Chat).
In this setting, we control (i)~the input context length by changing the number of documents in the input context (akin to retrieving more or less documents in retrieval-augmented generation), and (ii)~control the position of the relevant information within the input context by changing the order of the documents to place the relevant document at the beginning, middle or end of the context.

We find that changing the position of relevant information in the input context can substantially affect model performance, indicating that current language models do not robustly access and use information in long input contexts.
Furthermore, we observe a distinctive U-shaped performance curve (Figure~\ref{fig:figure1}); language model performance is highest when relevant information occurs at the very beginning (primacy bias) or end of its input context (recency bias), and performance significantly degrades when models must access and use information in the middle of their input context (\S\ref{sec:qa_results}).
For example, when relevant information is placed in the middle of its input context, \gptturbo's performance on the multi-document question task is lower than its performance when predicting \emph{without any documents} (i.e., the closed-book setting; 56.1\%).
Furthermore, we find that models often have identical performance to their extended-context counterparts, indicating that extended-context models are not necessarily better at using their input context (\S\ref{sec:qa_results}).

Given that language models struggle to retrieve and use relevant information in the multi-document question answering task, to what extent can language models even \emph{retrieve} from their input contexts? We study this question with a synthetic key-value retrieval task, which is designed to be a minimal testbed for the basic ability to retrieve matching tokens from the input context. In this task, models are given a collection of JSON-formatted key-value pairs and must return the value associated with a specific key. Similar to the multi-document QA task, the key-value retrieval task admits controlled changes to the input context length (adding more key-value pairs) and the position of relevant information.
Although some models perform the synthetic key-value retrieval task perfectly, other models struggle to simply retrieve matching tokens that occur in the middle of their input context and continue to exhibit a U-shaped performance curve.

To better understand why language models struggle to robustly access and use information in their input contexts, we study the role of model architecture (decoder-only vs. encoder-decoder), query-aware contextualization, and instruction fine-tuning (\S\ref{sec:why_u_shape}). We find that:
\begin{itemize}
\itemsep0em
\item Encoder-decoder models are relatively robust to changes in the position of relevant information within their input context, but only when evaluated on sequences within its training-time sequence length. When evaluated on sequences longer than those seen during training, we observe a U-shaped performance curve (\S\ref{sec:architecture}).
\item Query-aware contextualization (placing the query before \emph{and} after the documents or key-value pairs) enables near-perfect performance on the synthetic key-value task, but minimally changes trends in multi-document QA (\S\ref{sec:pre_conditioning}).
\item Even base language models (i.e., without instruction fine-tuning) show a U-shaped performance curve as we vary the position of relevant information in the input context.
\end{itemize}

Our results indicate that prompting language models with longer input contexts is a trade-off---providing the language model with more information may help it perform the downstream task, but it also increases the amount of content that the model must reason over, potentially decreasing accuracy.
To better understand this trade-off in practice, we perform a case study with retriever-reader models on open-domain question answering (\S\ref{sec:odqa_case_study}). In contrast to our controlled multi-document QA task, where the context always contains exactly \emph{one} document that answers the question, none or many of the top $k$ documents may contain the answer in the open-domain QA setting.
When retrieving from Wikipedia to answer queries from NaturalQuestions-Open, we find that model performance saturates long before retriever recall saturates, indicating that current models fail to effectively use additional retrieved documents---using 50 documents instead of 20 retrieved documents only marginally improves performance ($\sim$1.5\% for \gptturbo and $\sim$1\% for claude-1.3).

Our analysis provides a better understanding of how language models use their input context and introduces new evaluation protocols for future long-context models; to claim that a language model can robustly use information within long input contexts, it is necessary to show that its performance is minimally affected by the position of the relevant information in the input context (e.g., minimal difference in best- and worst-case performance).
To facilitate further work on understanding and improving how language models use their input context, we release our code and evaluation data.\footnote{\href{https://nelsonliu.me/papers/lost-in-the-middle}{nelsonliu.me/papers/lost-in-the-middle}}

\begin{figure*}[t]
\centering
\includegraphics[width=0.9\textwidth]{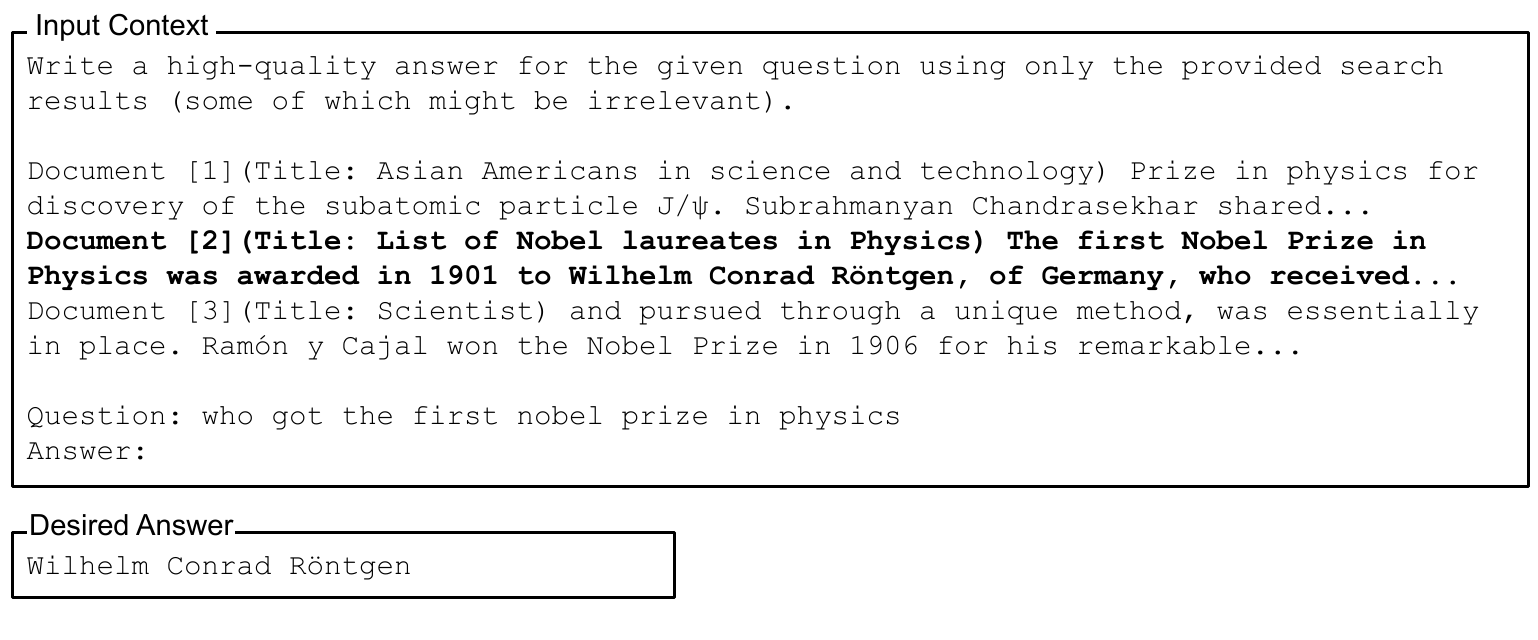}
\caption{Example of the multi-document question answering task, with an input context and the desired model answer. The document containing the answer is bolded within the input context here for clarity.
}\label{fig:qa_example}
\end{figure*}

\begin{figure}[t]
\centering
\includegraphics[width=0.9\columnwidth]{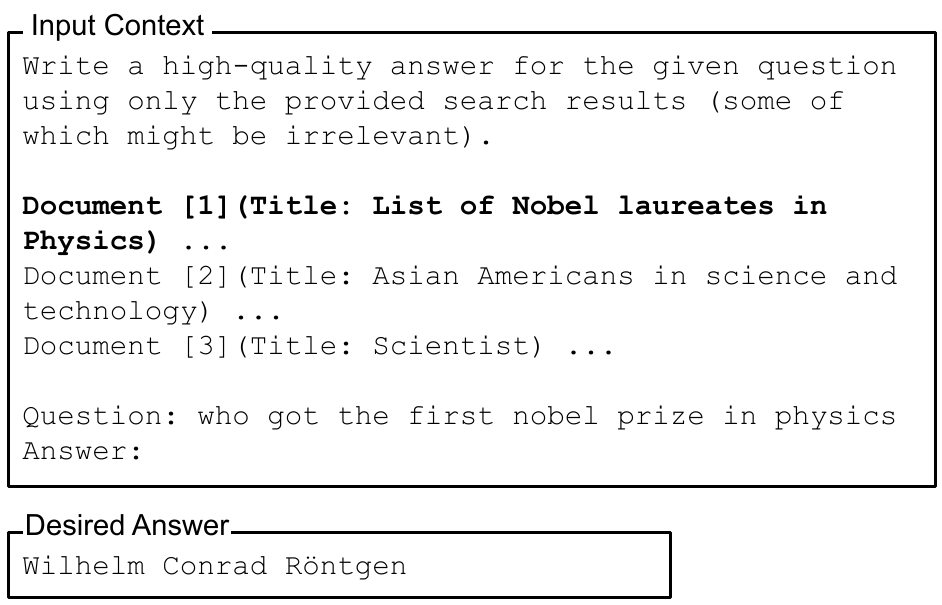}
\caption{Modulating the position of relevant information within the input context for the multi-document question answering example presented in Figure~\ref{fig:qa_example}. Re-ordering the documents in the input context does not affect the desired output.
}\label{fig:qa_changing_position}
\end{figure}

\begin{figure}[t]
\centering
\includegraphics[width=0.9\columnwidth]{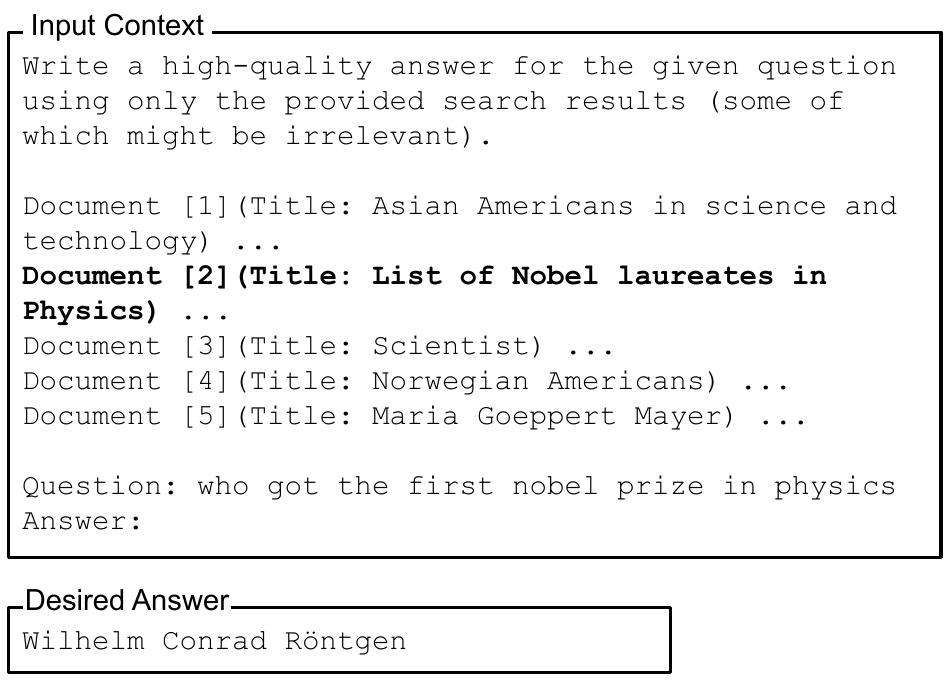}
\caption{Modulating the input context length of the multi-document question answering example presented in Figure~\ref{fig:qa_example}. Adding documents that do not contain the answer increases the length of the input context, but does not affect the desired output.
}\label{fig:qa_changing_length}
\end{figure}

\section{Multi-Document Question Answering}\label{sec:qa}

Our goal is to better understand how language models use their input context.
To this end, we analyze model performance on multi-document question answering, which requires models to find relevant information within an input context and use it to answer the question.
In particular, we make controlled changes to the length of the input context and the position of the relevant information and measure changes in task performance.

\subsection{Experimental Setup}

In the multi-document question answering task, the model inputs are (i)~a question to answer and (ii)~$k$ documents (e.g., passages from Wikipedia), where \emph{exactly one} of the documents contains the answer to the question and $k - 1$ ``distractor'' documents do not. This task requires the model to access the document that contains the answer within its input context and use it to answer the question.
Figure~\ref{fig:qa_example} presents an example.

\begin{figure*}[t]
\centering
\includegraphics[width=\textwidth]{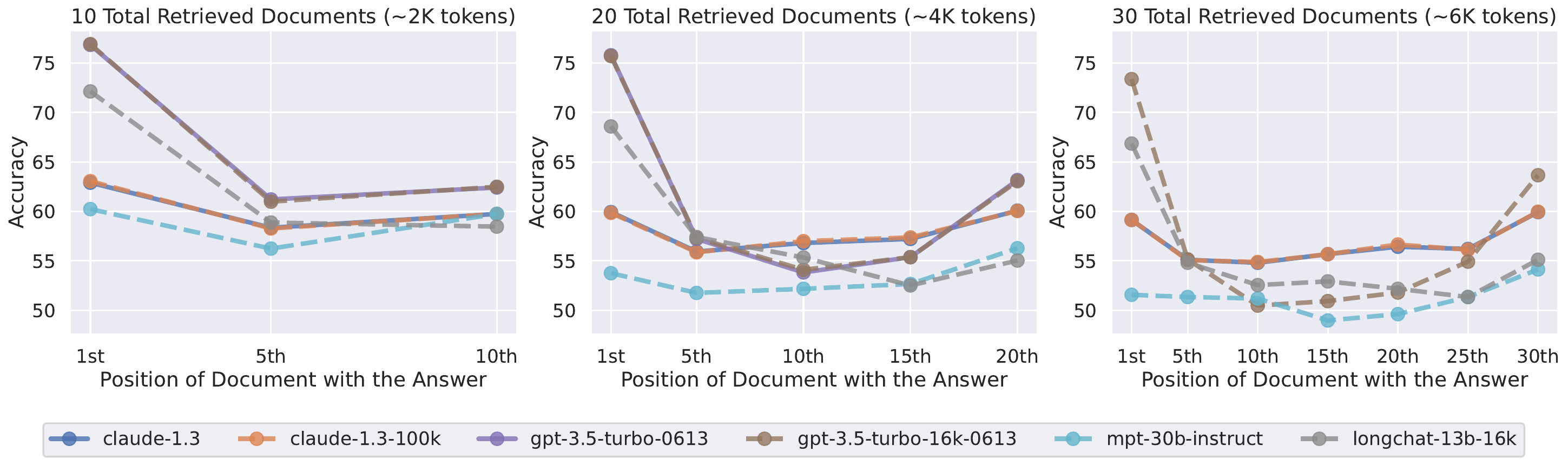}
\caption{The effect of changing the position of relevant information (document containing the answer) on multi-document question answering performance. Lower positions are closer to the start of the input context. Performance is highest when relevant information occurs at the very start or end of the context, and rapidly degrades when models must reason over information in the middle of their input context.
}\label{fig:qa_results}
\end{figure*}

We instantiate this task with data from NaturalQuestions-Open \citep{lee-etal-2019-latent,kwiatkowski-etal-2019-natural}, which contains historical queries issued to the Google search engine, coupled with human-annotated answers extracted from Wikipedia.
In particular, we take the 2655 queries where the annotated long answer is a paragraph (as opposed to a list or a table).
We use passages (chunks of at most 100 tokens) from Wikipedia as documents within our input contexts.
For each of the queries, we need a document that contains the answer and $k-1$ distractor documents that do not contain the answer.
To obtain a document that answers the question, we use the Wikipedia paragraph that contains the answer from the NaturalQuestions annotations.

To collect $k-1$ distractor documents that do not contain the answer, we use a retrieval system (Contriever, fine-tuned on MS-MARCO; \citealp{izacard2021contriever}) to retrieve the $k-1$ Wikipedia chunks that are most relevant to the query and do not contain any of the NaturalQuestions-annotated answers.\footnote{Ambiguity in NaturalQuestions-Open means that a small number of distractor passages may contain a reasonable answer. We additionally run experiments on subset of unambiguous questions, finding similar results and conclusions; see Appendix~\ref{sec:ambiguity}.}\footnote{We also explored using random documents as distractors, see Appendix~\ref{sec:random_distractors} for more details.}
In the input context, the distractor documents are presented in order of decreasing relevance.\footnote{Since there might be a prior over ``search results'' appearing in ranked order, we explored randomly ordering the $k-1$ distractor documents and mentioning that the documents are randomly ordered in the task description, but found the same trends. See Appendix~\ref{sec:qa_random_order} for more details.}

To modulate the position of relevant information within the input context, we adjust the order of the documents to change the position of the document that contains the answer (Figure~\ref{fig:qa_changing_position}).
To modulate the input context length in this task, we increase or decrease the number of retrieved documents that do not contain the answer (Figure~\ref{fig:qa_changing_length}).

Following \citet{kandpal2022large} and \citet{mallen2023trust}, we use accuracy as our primary evaluation metric, judging whether any of the correct answers (as taken from the NaturalQuestions annotations) appear in the predicted output.

Our experimental setup is similar to the needle-in-a-haystack experiments of \citet{ivgi-etal-2023-efficient}, who compare question answering performance when the relevant paragraph is placed (i)~at the beginning of the input or (ii)~a random position within the input. They find that encoder-decoder models have significantly higher performance when relevant information is placed at the start of the input context. In contrast, we study finer-grained changes in the position of relevant information.

\subsection{Models}\label{sec:models}

We analyze several state-of-the-art open and closed language models. We use greedy decoding when generating outputs and leave exploration of other decoding methods to future work. We use a standard set of prompts for each model (Figure~\ref{fig:qa_example}).

\paragraph{Open models.} We experiment with \mptinstruct, which has a maximum context length of 8192 tokens. The model was initially pre-trained on 1 trillion tokens using 2048-token sequences, followed by an additional sequence length adaptation pre-training phase on 50 billion tokens using 8192-token sequences. \mptinstruct uses ALiBi \citep{press2022train} to represent positional information.
We also evaluate \longchat \citep{longchat2023}, which extends the LLaMA-13B \citep{touvron2023llama} context window from 2048 to 16384 tokens by using condensed rotary positional embeddings before fine-tuning with 16384-token sequences.

\paragraph{Closed models.} We use the OpenAI API to experiment with \gptturbo and \gptturboextended.\footnote{We use the \texttt{0613} OpenAI model versions.}
\gptturbo has a maximum context length of 4K tokens, and \gptturboextended is a version with an extended maximum context length of 16K tokens.
We evaluate \claude and \claudeextended with the Anthropic API; \claude has a maximum context length of 8K tokens, and \claudeextended has an extended context length of 100K tokens.
\footnote{We also evaluate \gptfour (8K) on a subset of multi-document QA experiments, finding similar results and trends as other models (though GPT-4 has higher absolute performance). Evaluating \gptfour on the full multi-document QA and key-value retrieval experiments would cost upwards of \$6000. See Appendix~\ref{sec:gpt4} for GPT-4 results and discussion.}

\subsection{Results and Discussion}\label{sec:qa_results}
We experiment with input contexts containing 10, 20, and 30 total documents.
Figure~\ref{fig:qa_results} presents multi-document question answering performance when varying the position of relevant information within the input context.
To contextualize model performance, we also evaluate on the closed-book and oracle settings (Table~\ref{tab:closedbook_and_oracle}).
In the closed-book setting, models are not given any documents in their input context, and must rely on their parametric memory to generate the correct answer.
On the other hand, in the oracle setting, language models are given the single document that contains the answer and must use it to answer the question.

\begin{table}[t]
    \footnotesize
    \centering
    \begin{tabular}{@{}lrr@{}}
\toprule
Model & Closed-Book & Oracle  \\
\midrule
    \longchat & 35.0\% & 83.4\% \\
    \mptinstruct & 31.5\% & 81.9\% \\
    \gptturbo & 56.1\% & 88.3\% \\
    \gptturboextended & 56.0\% & 88.6\% \\
    \claude & 48.3\% & 76.1\% \\
    \claudeextended & 48.2\% & 76.4\% \\
    \bottomrule
    \end{tabular}
    \caption{Closed-book and oracle accuracy of language models on the multi-document question answering task.}
    \label{tab:closedbook_and_oracle}
\end{table}

\paragraph{Model performance is highest when relevant information occurs at the beginning or end of its input context.}
As illustrated in Figure~\ref{fig:qa_results}, changing the position of relevant information in the input context leads to substantial decreases in model performance. In particular, we see a distinctive U-shaped performance curve---models are often much better at using relevant information that occurs at the very beginning (primacy bias) and very end of contexts (recency bias), and suffer degraded performance when forced to use information within the middle of its input context. For example, \gptturbo's multi-document QA performance can drop by more than 20\%---in the worst case, performance in 20- and 30-document settings is lower than performance without \emph{any} input documents (i.e., closed-book performance; 56.1\%).
These results indicate that current models cannot effectively reason over their entire context window when prompted for downstream tasks.

\begin{figure*}[t]
\centering
\includegraphics[width=0.9\textwidth]{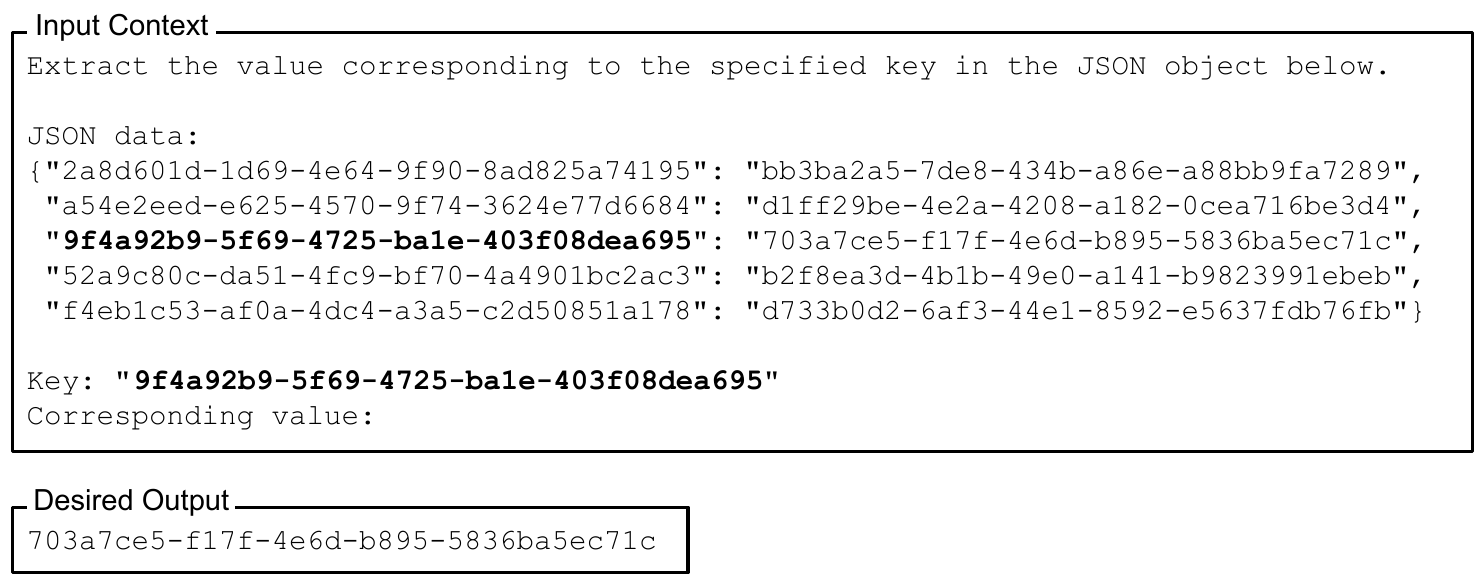}
\caption{Example of the key-value retrieval task, with an input context and the desired model output. Given a key, the goal is to return the associated value. All keys and values are 128-bit UUIDs. The relevant key-value pair for answering the query is bolded here within the input context for clarity.
}\label{fig:kv_retrieval_example}
\end{figure*}

\paragraph{Extended-context models are not necessarily better at using input context.}
When the input context fits in the context window of both a model and its extended-context counterpart, we see that performance between them is nearly identical. For example, the 10- and 20-document settings both fit in the context window of \gptturbo and \gptturboextended, and we observe that their performance as a function of position of relative information is nearly superimposed (solid purple and dashed brown series in Figure~\ref{fig:qa_results}). These results indicate that extended-context models are not necessarily better than their non-extended counterparts at using their input context.

\section{How Well Can Language Models Retrieve From Input Contexts?}\label{sec:kv}

Given that language models struggle to retrieve and use information from the middle of their input contexts in the multi-document question answering task, to what extent can they simply \emph{retrieve} from input contexts?
We study this question with a synthetic key-value retrieval task, which is designed to provide a minimal testbed for the basic ability to retrieve matching tokens from an input context.

\begin{figure*}[t]
\centering
\includegraphics[width=\textwidth]{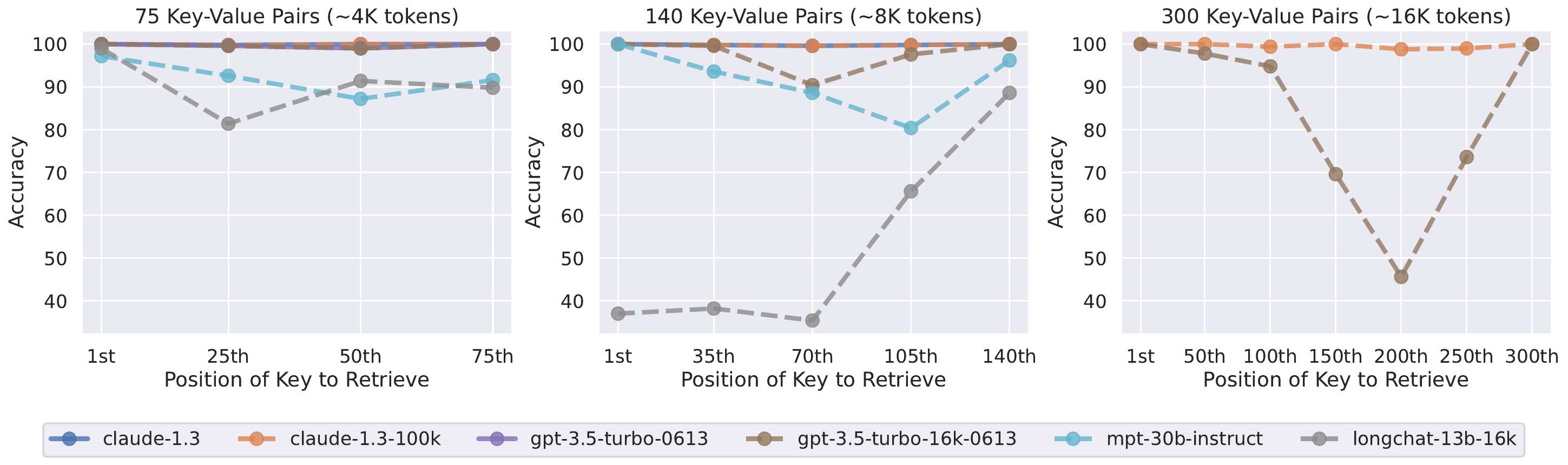}
\caption{The effect of changing the input context length and the position of relevant information on key-value retrieval performance. Lower positions are closer to the start of the input context. Although some models show perfect accuracy on this synthetic task (e.g., \claude{} and \claudeextended{}), we see again that performance is often highest when relevant information is occurs at the very start or end of the context, and rapidly degrades when models must retrieve from the middle of the input context.
}\label{fig:kv_results}
\end{figure*}

\subsection{Experimental Setup}

In our synthetic key-value retrieval task, the inputs are (i)~a string-serialized JSON object with $k$ key-value pairs, where each of the keys and values are unique, randomly-generated UUIDs and (ii)~a key within the aforementioned JSON object.
The goal is to return the value associated with the specified key.
Thus, each JSON object contains one relevant key-value pair (where the value is to be returned), and $k-1$ irrelevant ``distractor'' key-value pairs.
Figure~\ref{fig:kv_retrieval_example} provides an example input context and its corresponding desired output.
We again measure accuracy by evaluating whether the correct value appears in the predicted output.

Our synthetic key-value retrieval task shares similar goals with the Little Retrieval Test of \citet{littleretrievaltest} and the fine-grained line retrieval task of \citet{longchat2023}, but we explicitly seek to distill and simplify the task by removing as much natural language semantics as possible (using random UUIDs instead), since language features may present potential confounders.
For example, Transformer language models may have varying sensitivity to different linguistic features in their input \citep{oconnor-andreas-2021-context}.

To modulate the position of relevant information within the input context, we change the position of the key to retrieve within the serialized JSON object.
To modulate the input context length, we change the number of input JSON key-value pairs $k$ by adding or removing random keys, changing the number of distractor key-value pairs.

\subsection{Results and Discussion}\label{sec:kv_results}

We experiment with input contexts containing 75, 140, and 300 key-value pairs (500 examples each). We use the same set of models as the multi-document question answering experiments, see \S\ref{sec:models} for more details.

Figure~\ref{fig:kv_results} presents key-value retrieval performance. \claude{} and \claudeextended{} do nearly perfectly on all evaluated input context lengths, but other models struggle, especially when contexts have 140 or 300 key-value pairs---although the synthetic key-value retrieval task only requires identifying exact match within the input context, not all models achieve high performance.

Similar to our multi-document QA results, \gptturbo{}, \gptturboextended{}, and \mptinstruct{} have the lowest performance when they must access key-value pairs in the middle of their input context.
\longchat exhibits a different trend in the 140 key-value setting; we qualitatively observe that when relevant information is placed at the start of the input context, \longchat tends to generate code to retrieve the key, rather than outputting the value directly.

\section{Why Are Language Models Not Robust to Changes in the Position of Relevant Information?}\label{sec:why_u_shape}

Our multi-document question answering and key-value retrieval results show that language models struggle to robustly access and use information in long input contexts, since performance degrades significantly when changing the position of relevant information. To better understand why, we perform some preliminary investigations into the role of model architecture (decoder-only vs. encoder-decoder), query-aware contextualization, and instruction fine-tuning.

\begin{figure*}[t]
\centering
\includegraphics[width=\textwidth]{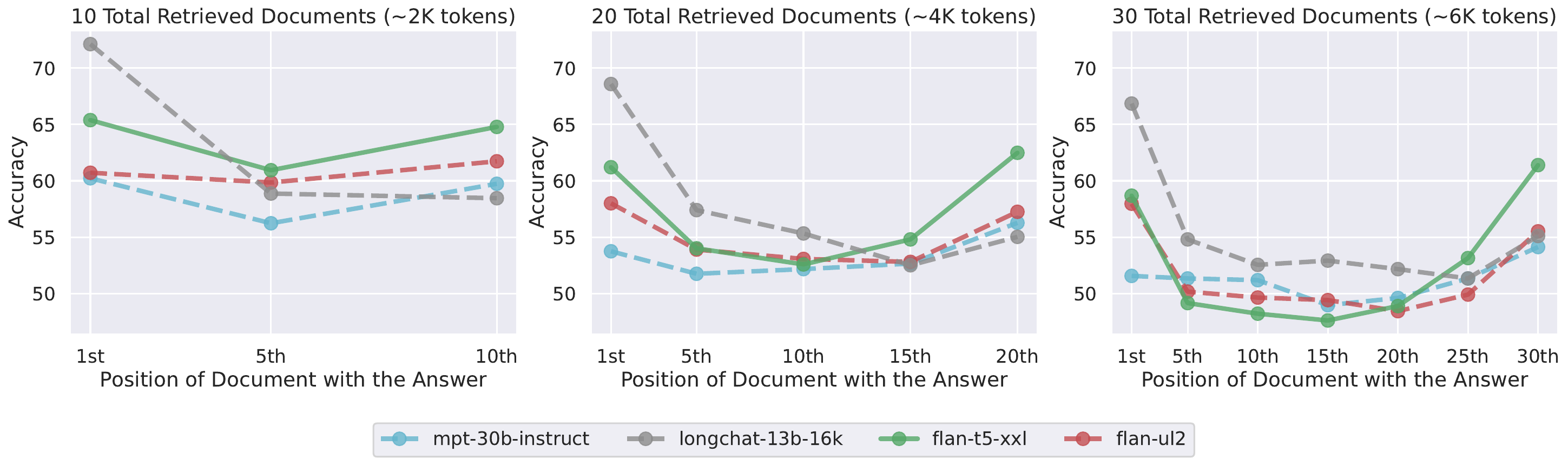}
\caption{When encoder-decoder models (\flanultwo and \flantfive) evaluated on sequences that are \emph{shorter} than their encoder's training-time maximum sequence length (2048 and 512 tokens, respectively), they are relatively robust to changes in the position of relevant information within their input context (left subplot).
In contrast, when these models are evaluated on sequences \emph{longer} than those seen during training (center and right subplots), we observe a U-shaped performance curve---performance is higher when relevant information occurs at the beginning or end of the input context, as opposed to the middle of the input context.
}\label{fig:qa_model_architecture}
\end{figure*}

\subsection{Effect of Model Architecture}\label{sec:architecture}

The open models we evaluated are all decoder-only models---at each timestep, they may only attend to prior tokens.
To better understand the potential effects of model architecture on how language model use context, we compare decoder-only and encoder-decoder language models.

We experiment with \flantfive \citep{JMLR:v21:20-074,chung2022scaling} and \flanultwo \citep{tay2023ul2}. Flan-T5-XXL is trained with a sequences of 512 tokens (encoder and decoder). Flan-UL2 is initially trained with sequences of 512 tokens (encoder and decoder), but is then pre-trained for an extra 100K steps with 1024 tokens (encoder and decoder) before instruction fine-tuning on sequences with 2048 tokens in the encoder and 512 tokens in the decoder.
However, since these models use relative positional embeddings, they can (in principle) extrapolate beyond these maximum context lengths; \citet{shaham2023zeroscrolls} find that both models can perform well with sequences of up to 8K tokens.

Figure~\ref{fig:qa_model_architecture} compares the performance of decoder-only and encoder-decoder models. When Flan-UL2 is evaluated on sequences within its 2048-token training-time context window (Figure~\ref{fig:qa_model_architecture}; left subplot), its performance is relatively robust to changes in the position of relevant information within the input context (1.9\% absolute difference between best- and worst-case performance). When evaluated on settings with sequences longer than 2048 tokens (Figure~\ref{fig:qa_model_architecture}; center and right), Flan-UL2 performance begins to degrade when relevant information is placed in the middle.
Flan-T5-XXL shows a similar trend, where longer input contexts result in a greater performance degradation when placing relevant information in the middle of the input context.
We hypothesize that encoder-decoder models may make better use of their context windows because their bidirectional encoder allows processing each document in the context of future documents, potentially improving relative importance estimation between documents.

\begin{figure}[t]
\centering
\includegraphics[width=0.9\columnwidth]{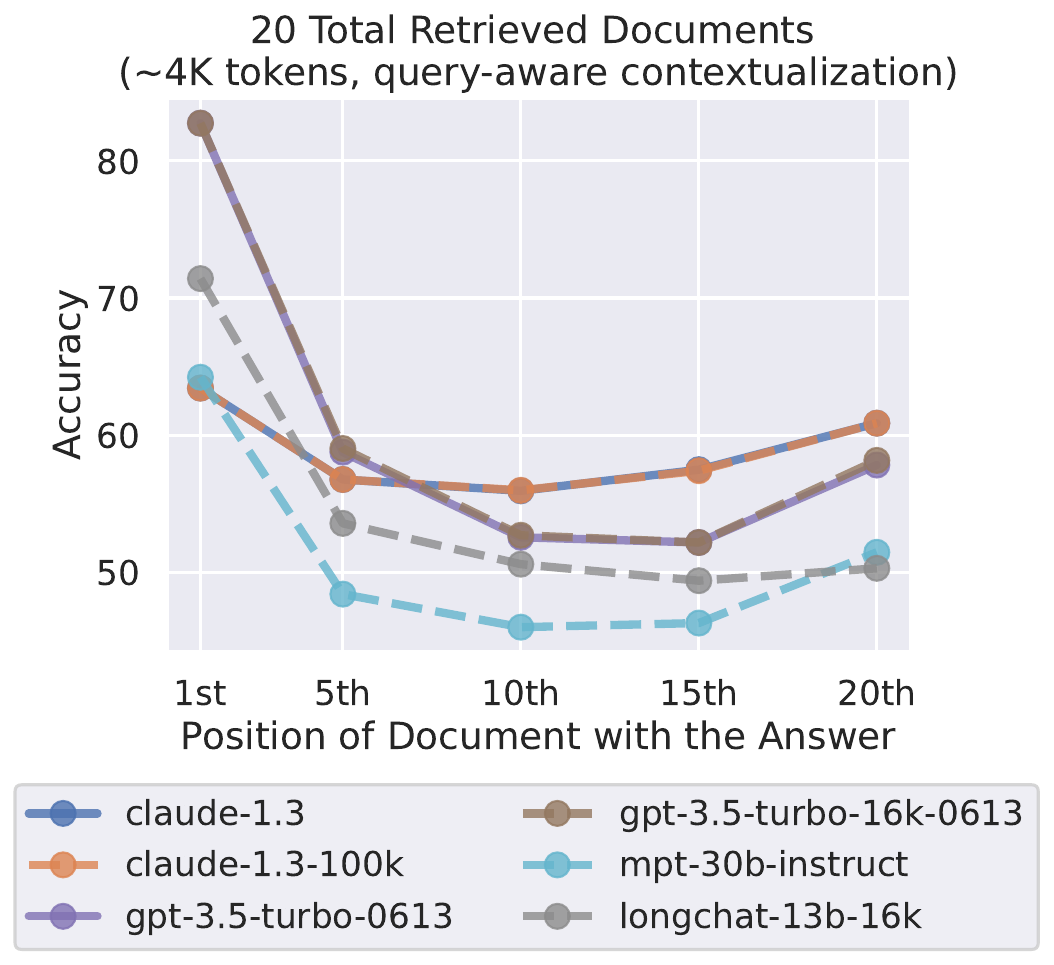}
\caption{Query-aware contextualization (placing the query before \emph{and} after the documents) does not substantially improve robustness of language models to changing the position of relevant information in multi-document QA; performance slightly increases when relevant information occurs at the very beginning, but otherwise slightly decreases.
}\label{fig:qa_query_preconditioning}
\end{figure}

\subsection{Effect of Query-Aware Contextualization}\label{sec:pre_conditioning}

Our multi-document QA and key-value retrieval experiments place the query (i.e., question to answer or key to retrieve) after the data to process (i.e., the documents or the key-value pairs). As a result, decoder-only models cannot attend to query tokens when contextualizing documents or key-value pairs, since the query only appears at the end of the prompt and decoder-only models can only attend to prior tokens at each timestep. In contrast, encoder-decoder models (which seem more robust to changes in the position of relevant information; \S\ref{sec:architecture}) use a bidirectional encoder to contextualize input contexts---can we use this observation to improve decoder-only models by placing the query before \emph{and} after the data, enabling query-aware contextualization of documents (or key-value pairs)?

We find that query-aware contextualization dramatically improves performance on the key-value retrieval task---all models achieve near-perfect performance on the 75, 140, and 300 key-value pair settings. For example, \gptturboextended with query-aware contextualization achieves perfect performance when evaluated with 300 key-value pairs.

In contrast, without query-aware contextualization, the worst-case performance is 45.6\% (Figure~\ref{fig:kv_results}).
Despite the significant impact on key-value retrieval performance, query-aware contextualization minimally affects performance trends in the multi-document question answering task (Figure~\ref{fig:qa_query_preconditioning}); it slightly improves performance when the relevant information is located at the very beginning of the input context, but slightly decreases performance in other settings.

\subsection{Effect of Instruction Fine-Tuning}\label{sec:instruction_tuning}

The models we evaluated are all instruction fine-tuned---after their initial pre-training, they undergo supervised fine-tuning on a dataset of instructions and responses.
The task specification and/or instruction is commonly placed at the beginning of the input context in supervised instruction fine-tuning data, which might lead instruction fine-tuned language models to place more weight on the start of the input context.
To better understand the potential effects of instruction fine-tuning on how language models use long input contexts, we compare the multi-document question answering performance of \mptinstruct against its base model (i.e., before instruction fine-tuning) \mpt. We use the same experimental setup as \S\ref{sec:qa}.

\begin{figure}[t]
\centering
\includegraphics[width=0.9\columnwidth]{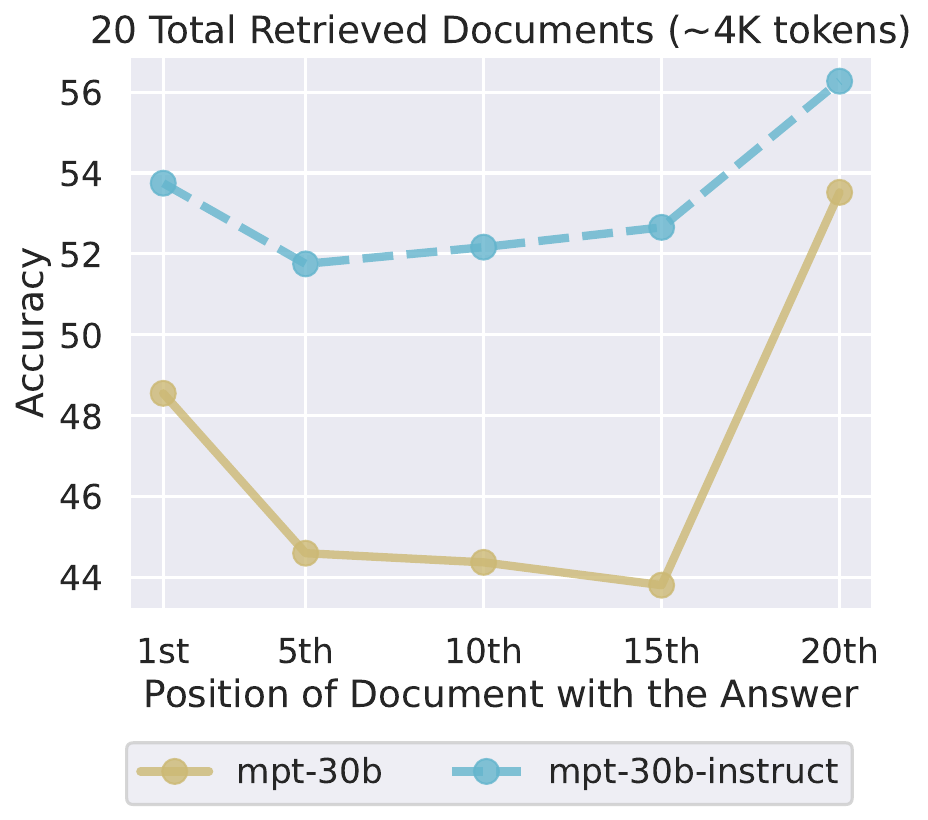}
\caption{Multi-document QA performance of \mptinstruct compared against its base model (i.e., before instruction fine-tuning) \mpt.
Both models have a U-shaped performance curve, where performance is much higher when relevant information occurs at the start or end of the input context, indicating that the instruction fine-tuning process itself is not necessarily responsible for these performance trends.
}\label{fig:qa_base_vs_instruction_tuned}
\end{figure}

Figure~\ref{fig:qa_base_vs_instruction_tuned} compares the multi-document QA performance of \mpt and \mptinstruct as a function of the position of the relevant information in the input context. Surprisingly, we see that both \mpt and \mptinstruct exhibit a U-shaped performance curve, where performance is highest when relevant information occurs at the very beginning or very end of the context. Although the absolute performance of \mptinstruct is uniformly higher than that of \mpt, their overall performance trends are similar. We also observe that instruction fine-tuning slightly reduces the worst-case performance disparity from nearly 10\% between the base model best- and worst-case performance to around 4\%.

These observations complement prior work, which found that non-instruction fine-tuned language models are biased towards recent tokens (i.e., the end of the input context; \citealp{khandelwal-etal-2018-sharp,press-etal-2021-shortformer}).
This recency bias has been observed in past work when evaluating models on next-word prediction of contiguous text, a setting where language models minimally benefit from long-range information \citep{sun-etal-2021-long}.
In contrast, our results show that language models are capable of using longer-range information (i.e., the beginning of the input context) when prompted with instruction-formatted data. We hypothesize that non-instruction fine-tuned language models learn to use these long contexts from similarly-formatted data that may occur in Internet text seen during pre-training, e.g., StackOverflow questions and answers.

To better understand the effect of additional fine-tuning and model scale, we also experimented with Llama-2 models of varying sizes (7B, 13B, and 70B) with and without additional supervised fine-tuning and reinforcement learning from human feedback (Appendix~\ref{sec:llama2}). We find that the U-shaped performance curve only appears in sufficiently large language models (with or without additional fine-tuning)---the 7B Llama-2 models are solely recency biased, while the 13B and 70B models exhibit a U-shaped performance curve. In addition, we see that the Llama-2 supervised fine-tuning and reinforcement learning from human feedback procedure slightly mitigates the positional bias in smaller models (13B, akin to trends shown when comparing \mpt and \mptinstruct), but minimally affects trends on larger models (70B).

\section{Is More Context Is Always Better? A~Case Study With Open-Domain QA}\label{sec:odqa_case_study}

Our results indicate that prompting language models with longer input contexts is a trade-off---providing the language model with more information may help it perform the downstream task, but it also increases the amount of content that the model must reason over, potentially decreasing accuracy. Even if a language model can take in 16K tokens, is it actually beneficial to provide 16K tokens of context? The answer to this question is ultimately downstream task-specific since it depends on the marginal value of the added context and the model's ability to effectively use long input contexts, but we perform a case study with open-domain question answering on NaturalQuestions-Open to better understand this trade-off in existing language models.

We use language models in a standard retriever-reader setup. A retrieval system (Contriever, fine-tuned on MS-MARCO) takes an input query from NaturalQuestions-Open and returns the $k$ documents from Wikipedia with the highest relevance score. To condition language models on these retrieved documents, we simply include them in the prompt. We evaluate retriever recall and reader accuracy (whether any of the annotated answers appear in the predicted output) as a function of the number of retrieved documents $k$. We use a subset of NaturalQuestions-Open where the long answer is a paragraph (as opposed to a table or a list).

\begin{figure}[t]
\centering
\includegraphics[width=0.9\columnwidth]{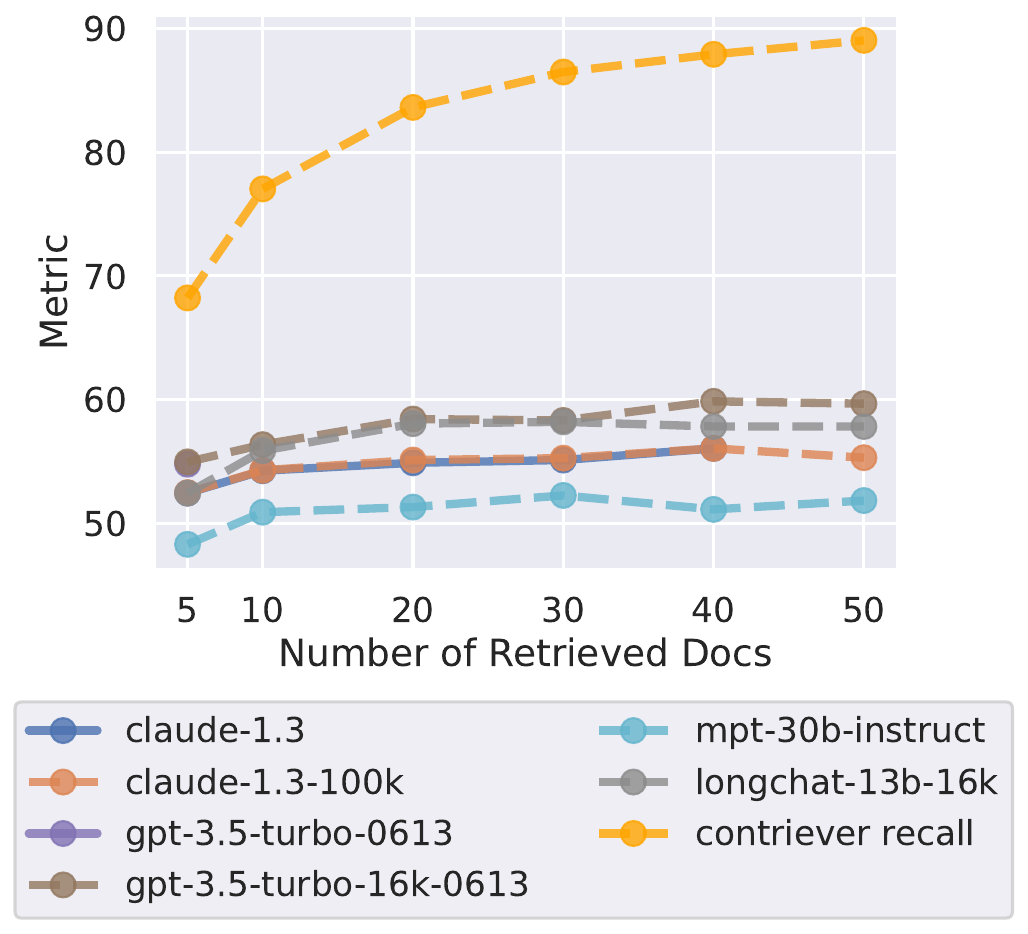}
\caption{Retriever recall and model performance as a function of the number of retrieved documents. Model performance saturates long before retriever recall, indicating that the models have difficulty making use of the extra retrieved documents.}\label{fig:odqa_results}
\end{figure}

Figure~\ref{fig:odqa_results} presents retriever recall and open-domain QA results. We see that reader model performance saturates long before retriever performance saturates, indicating that readers are not effectively using the extra context. Using more than 20 retrieved documents only marginally improves reader performance ($\sim$1.5\% for \gptturbo and $\sim$1\% for \claude), while significantly increasing the input context length (and thus latency and cost).
These results, coupled with the observation that models are often better at retrieving and using information at the start or end of the input contexts, suggest that effective reranking of retrieved documents (pushing relevant information closer to the start of the input context) or ranked list truncation (retrieving fewer documents when appropriate; \citealp{arampatzis2009stop}) may be promising directions for improving how language-model-based readers use retrieved context.

\section{Related Work}

\subsection{Long-Context Language Models}
There is much prior work in designing performant language models with cheaper scaling than Transformers in the context length.
Many lines of work pursue Transformer variants with attention modifications like recurrence \cite{dai-etal-2019-transformer}, factorizing attention into computationally less intensive approximations \cite{beltagy2020longformer,zaheer2020big}, or low-rank approximations \cite{Wang2020LinformerSW,peng2021random}.
\citet{dao2022flashattention} instead provide a faster exact attention by a carefully-crafted IO-aware CUDA kernel.
Separately, there are attempts to do away with attention entirely to remove quadratic sequence length complexity, often through convolution and/or linear RNNs, e.g., in RWKV \cite{PENG_RWKV-LM_2021}, S4 \cite{gu2022efficiently}, or Hyena \cite{poli2023hyena}.
Many prior efforts evaluate perplexity on a diverse web corpus as a proxy for the ability to process long contexts; this work shows that precise knowledge access on long contexts may be an added challenge.

\subsection{How Do Language Models Use Context?}
The pioneering work of \citet{khandelwal-etal-2018-sharp} showed that small LSTM language models make increasingly coarse use of longer-term context; \citet{sankar2019neural} found similar results in dialogue models.
In a similar vein, \citet{daniluk2017frustratingly} find that attentive LSTM language models tend to mainly use recent history.
\citet{petroni2020context} were among the first to demonstrate the potential of combining context from an information retrieval system with a pretrained language models for unsupervised question answering.
\citet{oconnor-andreas-2021-context} found that many information-destroying operations had marginal effects on Transformer LMs' predictions.
\citet{krishna2022rankgen} found that long-context neural generation in modestly-sized Transformer language models degenerates because models fail to properly condition on long context.
Finally, studying long-context models, \citet{sun-etal-2021-long} found that longer contexts improves prediction of only a few tokens, an empirical finding consistent with the theory of \citet{sharan2018prediction}, who showed that sequence distributions with bounded mutual information necessarily lead to marginal \textit{average} prediction benefits from increasingly long context.
\citet{qin-etal-2023-nlp} analyze how efficient Transformers perform on a variety of long-context downstream NLP tasks, finding that long-context transformers are recency-biased and do not effectively use long-range context.

\subsection{The Serial-Position Effect}
The U-shaped curve we observe in this work has a connection in psychology known as the \textit{serial-position effect} \cite{ebbinghaus1913memory,murdock1962serial}, that states that in free-association recall of elements from a list, humans tend to best remember the first and last elements of the list.
The serial-position effect plays a role in understanding how humans develop short- and long-term memory.
Observing a serial-position-like effect in language models is perhaps surprising, since the self-attention mechanisms underlying Transformer language models is technically equally capable of retrieving any token from their contexts.

\section{Conclusion}

We empirically study how language models use long input contexts via a series of controlled experiments.
We show that language model performance degrades significantly when changing the position of relevant information, indicating that models struggle to robustly access and use information in long input contexts.
In particular, performance is often lowest when models must use information in the middle of long input contexts.
We conduct a preliminary investigation of the role of (i)~model architecture, (ii)~query-aware contextualization, and (iii)~instruction fine-tuning to better understand how they affect how language models use context.
Finally, we conclude with a practical case study of open-domain question answering, finding that the performance of language model readers saturates far before retriever recall.
Our results and analysis provide a better understanding of how language models use their input context and provides new evaluation protocols for future long-context models.

\section*{Acknowledgments}

We would like to thank Luke Zettlemoyer, who served as our TACL action editor, and the the anonymous reviewers for their comments and feedback.
We also thank Claudiu Leoveanu-Condrei, Megan Leszczynski, Dmytro Okhonko, Maithra Raghu, Eric Wallace and Sang Michael Xie for feedback and discussions that helped improve this work.
Further, we are grateful to Sewon Min for her help with the AmbigQA dataset.
This work was supported by the Stanford Center for Research on Foundation Models (CRFM), by OpenAI via an API credits grant to the Stanford CRFM, and by Anthropic via the Claude academic access program.

\bibliography{custom}
\bibliographystyle{acl_natbib}

\appendix

\section{Ambiguity in Multi-Document QA Distractor Documents}\label{sec:ambiguity}

Following past work on NaturalQuestions-Open \citep[\emph{inter alia}]{izacard2021contriever,izacard-grave-2021-leveraging}, we use a Wikipedia dump from late 2018 as our retrieval corpus. However, this standard Wikipedia dump has a small amount of temporal mismatch with the NaturalQuestions annotations.

For example, consider the question ``what nfl team does robert griffin iii play for''. The NaturalQuestions annotated answer is ``currently a free agent''. However, the Wikipedia retrieval corpus contains the information that he plays for the ``Baltimore Ravens'', since he was released from the team between the Wikipedia dump's timestamp and the NaturalQuestions annotation process.

\begin{figure}
\centering
\includegraphics[width=\columnwidth]{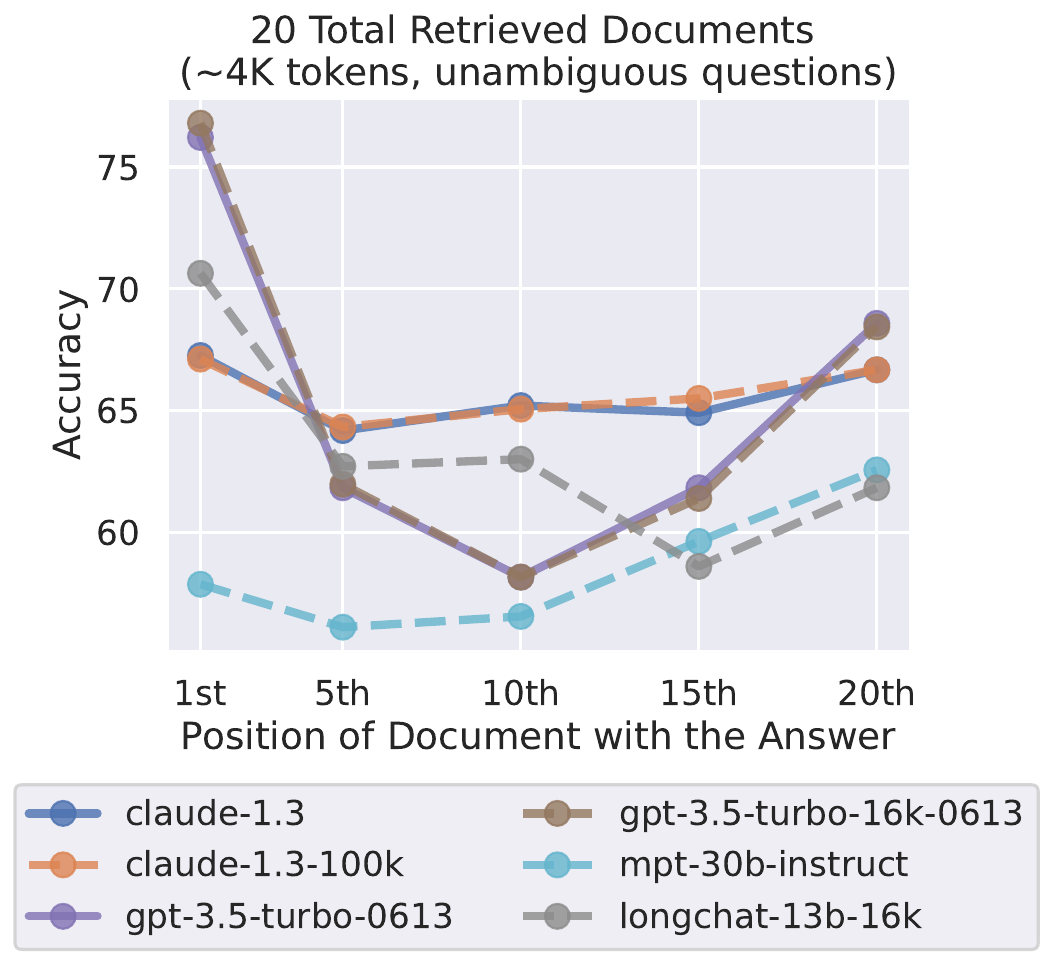}
\caption{Language model performance on a unambiguous subset of questions.
}\label{fig:qa_nonambiguous}
\end{figure}

We use the ambiguity annotations of \citet{min-etal-2020-ambigqa} to create a subset unambiguous questions. Experiments on this unambiguous subset of the data show similar results and conclusions as the experiments on the full questions collection (Figure~\ref{fig:qa_nonambiguous}).

\section{Random Distractors in Multi-Document QA}\label{sec:random_distractors}

\begin{figure}
\centering
\includegraphics[width=\columnwidth]{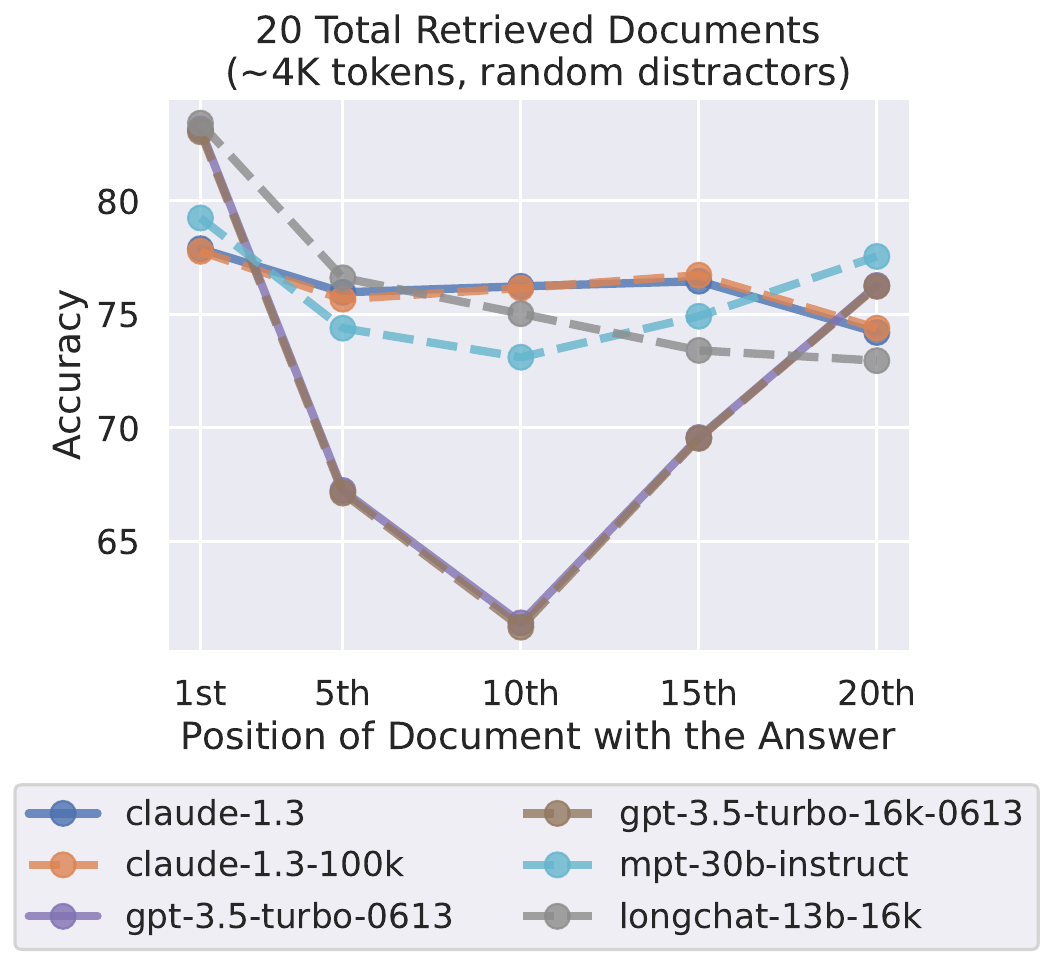}
\caption{Language model performance on multi-document QA when using random distractors, rather than retrieved distractors.
}\label{fig:qa_random_distractors}
\end{figure}

We also run multi-document question answering experiments with random Wikipedia documents as distractors, which allows us to ablate the impact of retrieved distractors (hard negatives).
Note that in this setting, the the document containing the answer can often be identified with simple heuristics
(e.g., lexical overlap with the query).
Figure~\ref{fig:qa_random_distractors} presents the results of this experiment.
Although all models have higher absolute accuracy in this setting, they
surprisingly still struggle to reason over their entire input context, indicating that their performance degradation is not solely due to an inability to identify relevant documents.

\section{Randomizing Distractor Order in Multi-Document QA}
\label{sec:qa_random_order}

Our prompt instructs the language model to use the provided search results to answer the question. There may be a prior in the pre-training or instruction fine-tuning data to treat search results as sorted by decreasing relevance (i.e., the documents near the beginning of the input context are more likely to be useful than those at the end). To validate that our conclusions are not simply a byproduct of this bias, we run experiments with the modified instruction ``Write a high-quality answer for the given question using only the provided search results (some of which might be irrelevant). The search results are ordered randomly.''
In addition, we randomly shuffle the $k-1$ distractor documents.

\begin{figure}
\centering
\includegraphics[width=\columnwidth]{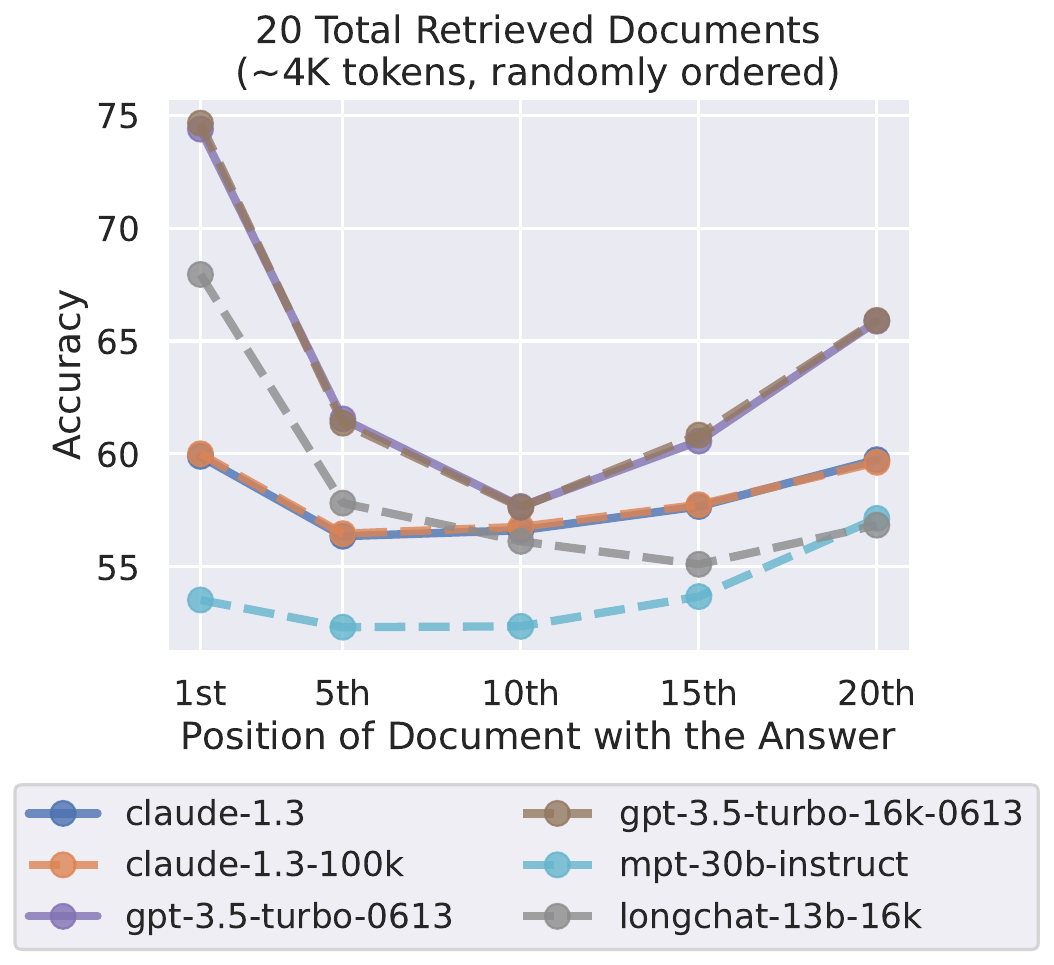}
\caption{Language model performance when randomizing the order of the distractors (rather than presenting them in order of decreasing relevance) and mentioning as such in the prompt.
}\label{fig:qa_random_order}
\end{figure}

Figure~\ref{fig:qa_random_order} presents the results of this experiment.
We continue to see a U-shaped performance curve, with performance degrading when language models must use information in the middle of their input contexts.
Comparing the results in \S\ref{sec:qa_results} with those when randomizing the distractor order and mentioning such in the prompt, we see that randomization slightly decreases performance when the relevant information is at the very beginning of the context, and slightly increases performance when using information in the middle and end of the context.

\section{\gptfour Performance}
\label{sec:gpt4}

\begin{figure}
\centering
\includegraphics[width=\columnwidth]{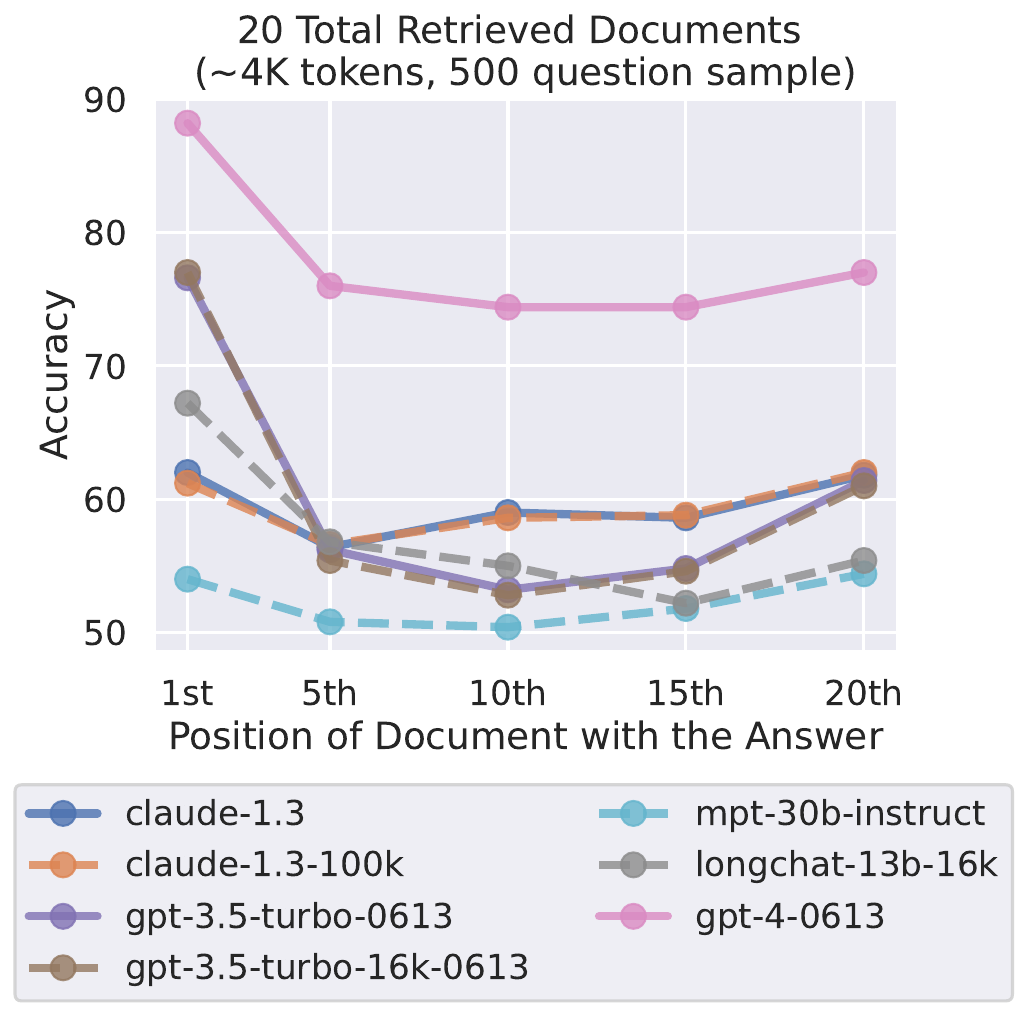}
\caption{Although \gptfour has higher absolute performance than other models, its performance still degrades when relevant information occurs in the middle of the input context.
}\label{fig:gpt4}
\end{figure}

We evaluate \gptfour (8K) on a subset of 500 random multi-document QA examples with 20 total documents in each input context (Figure~\ref{fig:gpt4}). GPT-4 achieves higher absolute performance than any other language model, but still shows a U-shaped performance curve---its performance is highest when relevant information occurs at the very start or end of the context, and performance degrades when it must use information in the middle of its input context.

\section{Llama-2 Performance}\label{sec:llama2}

\begin{figure}
\centering
\includegraphics[width=\columnwidth]{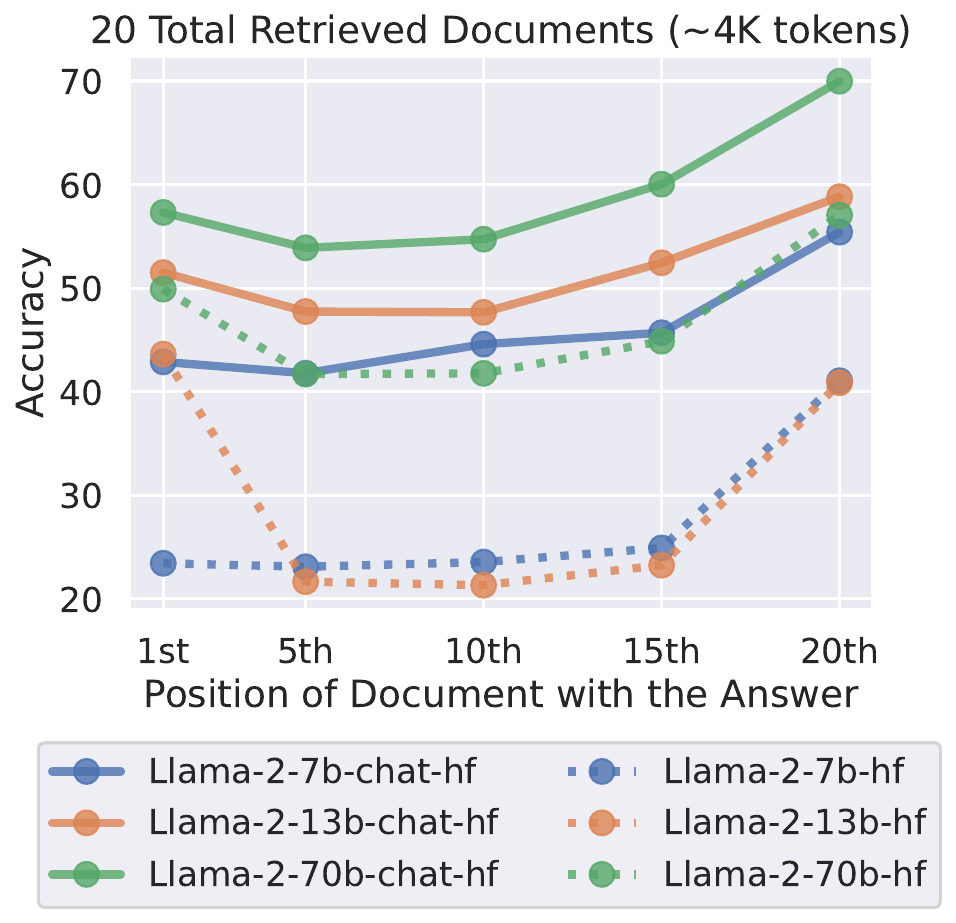}
\caption{Multi-document QA performance (20 total documents) of Llama-2 models of varying sizes (7B, 13B, 70B parameters), with and without additional supervised fine-tuning and reinforcement learning from human feedback (``\texttt{-chat-}'' models).}\label{fig:llama2}
\end{figure}

We evaluate Llama-2 \citep{touvron2023llama2} on multi-document QA with 20 total documents in each input context.
The Llama tokenizer produces longer sequences than the tokenizers for our previously-studied models, so we discard 20 examples (out of 2655) that exceed Llama-2's maximum context length of 4096 tokens.
We experiment with models of varying sizes (7B, 13B, and 70B parameters), with and without additional supervised fine-tuning and reinforcement learning from human feedback (``\texttt{-chat-}'' models). The results are presented in Figure~\ref{fig:llama2}.

Comparing Llama-2 models of varying sizes, we find that only the larger models (13B and 70B) exhibit the U-shaped performance curve (i.e., both primacy and recency bias)---the smallest Llama-2 models (7B) are solely recency-biased. Given these results, we hypothesize that prior work (e.g., \citealp{khandelwal-etal-2018-sharp,sun-etal-2021-long}) did not previously observe any primacy bias in language models because the models they studied were too small (less than 1B parameters).

Comparing between Llama-2 models with and without additional supervised fine-tuning and reinforcement learning from human feedback, we see that additional fine-tuning dramatically improves performance on the multi-document QA task.
The 7B models with and without additional fine-tuning show minimal primacy bias, and are largely recency-biased.
The 13B base model has a dramatic primacy and recency bias---there is a 20-point accuracy disparity between the best- and worst-case performance.
Applying additional fine-tuning to the 13B seems to slightly reduce this bias (10-point worst-case degradation), but the bias remains significant.
However, the 70B models with and without additional fine-tuning have largely similar trends (showing both primacy and recency bias), and additional fine-tuning minimally changes the positional bias severity.

\clearpage
\newpage

\onecolumn
\section{Token Counts} \label{sec:token_counts}

Table~\ref{tab:qa_oracle_closedbook_tokens}, Table~\ref{tab:qa_tokens}, and Table~\ref{tab:kv_tokens} present the average and maximum number of tokens in each of the input contexts for all experimental settings. Note that \mpt and \mptinstruct use the same tokenizer, \gptturbo and \gptturboextended use the same tokenizer, and \claude and \claudeextended use the same tokenizer. Furthermore, the \claude tokenizer is the same as the \gptturbo tokenizer, modulo some additional special tokens that do not appear in our data. As a result, the token counts for these two model families is the same in our experimental settings.

\begin{table*}[h]
    \centering
    \begin{tabular}{@{}lcccc@{}}
\toprule
&  \multicolumn{2}{c}{Closed-Book} &  \multicolumn{2}{c}{Oracle}\\
\cmidrule(l){2-3} \cmidrule(l){4-5}
& avg $\pm$ stdev & max & avg $\pm$ stdev & max\\
\midrule
    \longchat & 55.6 $\pm$ 2.7 & 70 & 219.7 $\pm$ 48.5 & 588 \\
    \mpt & 43.5 $\pm$ 2.2 & 58 & 187.9 $\pm$ 41.8 & 482 \\
    \gptturbo & 15.3 $\pm$ 2.2 & 29 & 156.0 $\pm$ 41.8 & 449 \\
    \claude & 15.3 $\pm$ 2.2 & 29 & 156.0 $\pm$ 41.8 & 449 \\
    \bottomrule
    \end{tabular}
    \caption{Token count statistics for each of the evaluated models on the closed-book and oracle multi-document question answering settings.}\label{tab:qa_oracle_closedbook_tokens}
\end{table*}

\begin{table*}[h]
    \centering
    \begin{tabular}{@{}lcccccc@{}}
\toprule
&  \multicolumn{2}{c}{10 docs} &  \multicolumn{2}{c}{20 docs} &  \multicolumn{2}{c}{30 docs}\\
\cmidrule(l){2-3} \cmidrule(l){4-5} \cmidrule(l){6-7}
& avg $\pm$ stdev & max & avg $\pm$ stdev & max & avg $\pm$ stdev & max \\
\midrule
    \longchat & 1749.9 $\pm$ 112.4 & 2511 & 3464.6 $\pm$ 202.3 & 4955 & 5181.9 $\pm$ 294.7 & 7729 \\
    \mpt & 1499.7 $\pm$ 88.5 & 1907 & 2962.4 $\pm$ 158.4 & 3730 & 4426.9 $\pm$ 230.5 & 5475 \\
    \gptturbo & 1475.6 $\pm$ 86.5 & 1960 & 2946.2 $\pm$ 155.1 & 3920 & 4419.2 $\pm$ 226.5 & 6101\\
    \claude & 1475.6 $\pm$ 86.5 & 1960 & 2946.2 $\pm$ 155.1 & 3920 & 4419.2 $\pm$ 226.5 & 6101\\
    \bottomrule
    \end{tabular}
    \caption{Token count statistics for each of the evaluated models on each of the document question answering settings.}\label{tab:qa_tokens}
\end{table*}

\begin{table*}[h]
    \centering
    \begin{tabular}{@{}lcccccc@{}}
\toprule
&  \multicolumn{2}{c}{75 KV pairs} &  \multicolumn{2}{c}{140 KV pairs} &  \multicolumn{2}{c}{300 KV pairs}\\
\cmidrule(l){2-3} \cmidrule(l){4-5} \cmidrule(l){6-7}
& avg $\pm$ stdev & max & avg $\pm$ stdev & max & avg $\pm$ stdev & max \\
\midrule
    \longchat & 5444.5 $\pm$ 19.1 & 5500 & 10072.4 $\pm$ 24.1 & 10139 & 21467.3 $\pm$ 35.9 & 21582 \\
    \mpt & 4110.5 $\pm$ 23.8 & 4187 & 7600.9 $\pm$ 31.1 & 7687 & 16192.4 $\pm$ 46.6 & 16319 \\
    \gptturbo & 3768.7 $\pm$ 25.6 & 3844 & 6992.8 $\pm$ 34.1 & 7088 & 14929.4 $\pm$ 50.7 & 15048\\
    \claude & 3768.7 $\pm$ 25.6 & 3844 & 6992.8 $\pm$ 34.1 & 7088 & 14929.4 $\pm$ 50.7 & 15048\\
    \bottomrule
    \end{tabular}
    \caption{Token count statistics for each of the evaluated models on each of the key-value (KV) retrieval settings.}\label{tab:kv_tokens}
\end{table*}

\clearpage
\newpage

\section{Full Multi-Document Question Answering Results}

This section tabulates model performance when evaluated on the multi-document QA task with varying numbers of documents (Figure~\ref{fig:qa_results}). ``Index $n$'' indicates performance when the document with the answer occurs at position $n + 1$, where lower indices are closer to the start of the input context. For example, index 0 refers to performance when the document with the answer is placed at the very start of the context (i.e., first amongst all documents).

\subsection{10 Total Retrieved Documents}

\begin{table}[h]
\centering
\begin{tabular}{lrrr}
\toprule
Model & Index 0 & Index 4 & Index 9 \\
\midrule
\claude & 62.9\% & 58.3\% & 59.7\% \\
\claudeextended & 63.1\% & 58.3\% & 59.7\% \\
\gptturbo & 76.8\% & 61.2\% & 62.4\% \\
\gptturboextended & 76.9\% & 61.0\% & 62.5\% \\
\mptinstruct & 60.2\% & 56.2\% & 59.7\% \\
\longchat & 72.1\% & 58.9\% & 58.5\% \\
\bottomrule
\end{tabular}
\caption{Model performance when evaluated on the multi-document QA task with 10 total retrieved documents.}
\end{table}

\subsection{20 Total Retrieved Documents}

\begin{table}[h]
\centering
\begin{tabular}{lrrrrr}
\toprule
Model & Index 0 & Index 4 & Index 9 & Index 14 & Index 19 \\
\midrule
\claude & 59.9\% & 55.9\% & 56.8\% & 57.2\% & 60.1\% \\
\claudeextended & 59.8\% & 55.9\% & 57.0\% & 57.4\% & 60.0\% \\
\gptturbo & 75.8\% & 57.2\% & 53.8\% & 55.4\% & 63.2\% \\
\gptturboextended & 75.7\% & 57.3\% & 54.1\% & 55.4\% & 63.1\% \\
\mptinstruct & 53.7\% & 51.8\% & 52.2\% & 52.7\% & 56.3\% \\
\longchat & 68.6\% & 57.4\% & 55.3\% & 52.5\% & 55.0\% \\
\bottomrule
\end{tabular}
\caption{Model performance when evaluated on the multi-document QA task with 20 total retrieved documents.}
\end{table}

\subsection{30 Total Retrieved Documents}

\begin{table}[h]
\centering
\begin{tabular}{lrrrrrrr}
\toprule
Model & Index 0 & Index 4 & Index 9 & Index 14 & Index 19 & Index 24 & Index 29 \\
\midrule
\claude & 59.1\% & 55.1\% & 54.8\% & 55.7\% & 56.4\% & 56.2\% & 59.9\% \\
\claudeextended & 59.1\% & 55.1\% & 54.9\% & 55.7\% & 56.6\% & 56.1\% & 60.0\% \\
\gptturboextended & 73.4\% & 55.1\% & 50.5\% & 50.9\% & 51.8\% & 54.9\% & 63.7\% \\
\mptinstruct & 51.6\% & 51.3\% & 51.2\% & 49.0\% & 49.6\% & 51.3\% & 54.1\% \\
\longchat & 66.9\% & 54.8\% & 52.5\% & 52.9\% & 52.2\% & 51.3\% & 55.1\% \\
\bottomrule
\end{tabular}
\caption{Model performance when evaluated on the multi-document QA task with 30 total retrieved documents.}
\end{table}

\end{document}